\documentclass[10pt,twocolumn,letterpaper]{article}

\usepackage[numbers]{natbib}

\usepackage[utf8]{inputenc} 
\usepackage[T1]{fontenc}    
\usepackage{url}            
\usepackage{booktabs}       
\usepackage{amsfonts}       
\usepackage{nicefrac}       
\usepackage{microtype}      
\usepackage{graphicx}
\usepackage{amsmath}
\usepackage{amssymb}
\usepackage{multirow}
\usepackage{verbatim}
\usepackage{xcolor}
\usepackage{float}
\usepackage{csquotes}
\usepackage[ruled]{algorithm2e}

\title{Unsupervised Multi-Target Domain Adaptation Through \\ Knowledge Distillation}

\author{%
  L.T. Nguyen-Meidine$^\alpha$, A. Belal$^\beta$, M. Kiran$^\alpha$, J. Dolz$^\alpha$, L-A. Blais-Morin$^\gamma$, E. Granger$^\alpha$   \\ 
$^\alpha$ {\textit{LIVIA, École de technologie supérieure, Montreal, Canada}}\\
$^\beta$ {\textit{Aligarh Muslim University, Aligarh, India}}\\
$^\gamma$ {\textit{Genetec Inc., Montreal Canada}}\\
  {\tt\small le-thanh.nguyen-meidine.1@ens.etsmtl.ca,}
  {\tt\small abelal@myamu.ac.in,}
  {\tt\small mkiran@livia.etsmtl.ca,} \\
  {\tt\small \{jose.dolz, eric.granger\}@etsmtl.ca,}
  {\tt\small lablaismorin@genetec.com}
}

\date{}
\begin{document}

\maketitle

\begin{abstract}
Unsupervised domain adaptation (UDA) seeks to alleviate the problem of domain shift between the distribution of unlabeled data from the target domain w.r.t. labeled data from the source domain. While the single-target UDA  scenario is well studied in the literature, Multi-Target Domain Adaptation (MTDA) remains largely unexplored despite its practical importance, e.g., in multi-camera video-surveillance applications.  
The MTDA problem can be addressed by adapting one specialized model per target domain, although this solution is too costly in many real-world applications. Blending multiple targets for MTDA has been proposed, yet this solution may lead to a reduction in model specificity and accuracy. 
In this paper, we propose a novel unsupervised MTDA approach to train a CNN that can generalize well across multiple target domains. Our Multi-Teacher MTDA (MT-MTDA) method relies on multi-teacher knowledge distillation (KD) to iteratively distill target domain knowledge from multiple teachers to a common student. The KD process is performed in a progressive manner, where the student is trained by each teacher on how to perform UDA for a specific target, instead of directly learning domain adapted features. Finally, instead of combining the knowledge from each teacher, MT-MTDA alternates between teachers that distill knowledge, thereby preserving the specificity of each target (teacher) when learning to adapt to the student. 
%
MT-MTDA is compared against state-of-the-art methods on several challenging UDA benchmarks, and empirical results show that our proposed model can provide a considerably higher level of accuracy across multiple target domains. Our code is available at: \url{https://github.com/LIVIAETS/MT-MTDA}.

\end{abstract}

\section{Introduction}

Deep Learning (DL) models, and in particular Convolutional Neural Networks (CNNs), have achieved state-of-the-art performance in many visual recognition applications such as image classification, detection and segmentation \cite{Goodfellow-et-al-2016}.  Despite their success, several factors limit their deployment in real-world industrial applications. Among these factors is the problem of domain shift, where the distribution of original training data (source domain) diverges w.r.t data from the operational environment (target domain). This problem often translates to a notable decline in performance once the DL model has been deployed in the target domain. 

To address this problem, DL models for domain adaptation have been proposed to align a discriminant source model with the target domain using data captured from the target domain \cite{ekladious2020, GRL, luo2018, wen2018}. In unsupervised domain adaptation (UDA), a large amount of unlabeled data is often assumed to have been collected from the target domain to avoid the costly task of annotating data. Currently, several conventional and DL models have been proposed for the single target domain adaptation (STDA) setting, using unlabeled data that is collected from a single target domain. These models rely on different approaches, ranging from the optimization of a statistical criterion to the integration of adversarial losses, in order to learn robust domain-invariant representations from source and target domain data. However, despite multi-target domain adaptation (MTDA) scenario, i.e. multiple unlabeled target domains, has many real-world applications, it remains virtually unexplored.  For instance, in video-surveillance applications, each camera of a distributed network corresponds to a different non-overlapping viewpoint (target domain). A DL model for person re-identification \cite{mekhazni2020unsupervised} should normally be adapted to multiple different camera viewpoints.

\begin{figure*}[!t]
    \centering
    \includegraphics[width=1\textwidth]{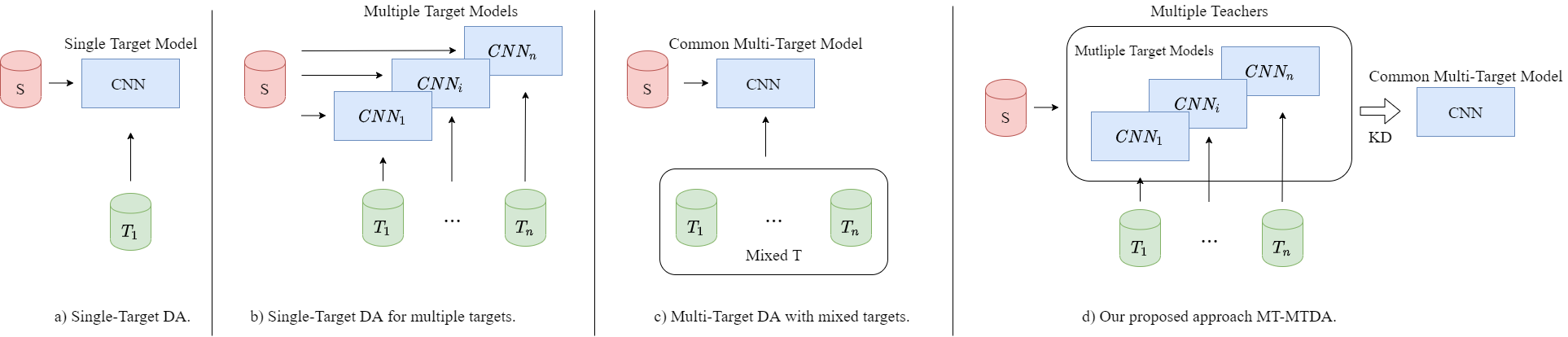}
    \caption{Illustration of different STDA and MTDA strategies for training CNNs across multiple target domains. $S$ is the labelled source dataset, while $T_i$ are the unlabelled target datasets for $i = 1, 2 ..., n$.}
    \label{fig:Survey_MTDA}
\end{figure*}

Extension of STDA techniques to the MTDA setting is not straightforward, and they may perform poorly on multiple target domains. Although MTDA problems can be solved by producing one model per target domain, this approach becomes costly and impractical in applications with a growing number of target domains. In such cases, a MTDA approach should ideally yield a common DL model that is compact and has been adapted to perform accurately across all target domains. To adapt a common multi-target DL model, one recent MTDA approach considers the problem of MTDA without domain labels, and proposes an approach to blend all the target domains together, which may lead to a reduction in model specificity and accuracy \cite{BlendMTDA}. While the current approach provides an interesting direction in adapting a common model to multiple target domains, we argue that directly adapting a model to multiple target domains can affect the performance since there are limitations on a model's capacity to learn and generalize in diverse target domains. Other works on MTDA have focused on the problem of unshared categories between target domains \cite{MTDA_No_shared}, nevertheless, this scenario has not been considered since it is outside the scope of this work. 


In this paper, a novel MTDA learning strategy referred to as Multi-Teacher MTDA (MT-MTDA) is proposed to train a common CNN to perform well across multiple target domains. Our strategy relies on knowledge distillation to efficiently transfer information from several different target domains, each one associated with a specialized teacher, to a single common multi-target model. Figures \ref{fig:Survey_MTDA}(a)-(c) illustrate the different MTDA strategies from literature, evolving from strategies that adapt a single CNN per target domain, to strategies that adapt a common CNN across all target domains. Our novel MT-MTDA approach (illustrated in Figure \ref{fig:Survey_MTDA}(d)) is inspired by a common education scenario, where each teacher is responsible for a single subject (i.e. target domain), and these teachers sequentially educate a student to learn all the subjects. 

In our MT-MTDA approach: (1) Since only the student performance is important after training, we can resort to complex architecture for the teacher model; (2) These complex teachers can provide a higher capacity to generalize toward a single target domain instead of having one model learning multiple target domains; (3) The student model learns compressed knowledge from teachers across target domains, instead of directly learning to generalize on multiple domains; and (4) MT-MTDA can benefit from different STDA algorithms since each teacher adapt to only one target. 



We also propose an efficient alternative for the fusion of knowledge from multiple teachers. State-of-the-art techniques for multi-teacher knowledge distillation rely on average fusion (sum operations) to directly combine the information derived by teachers \cite{KnowledgeAdaptation}. To preserve the specificity of individual teachers, we let our student model learn to adapt from each teacher separately and sequentially from teacher to teacher. We argue that having better  preservation  of target specificity leads to higher accuracy. 

Finally, the proposed MT-MTDA is compared extensively to state-of-the-art strategies on widely used UDA benchmarks (OfficeHome, Office31, and Digits-5), and show that MT-MTDA consistently achieves a high level of accuracy across multiple target domains with different backbone network architectures.

\vspace{-3mm}
\section{Related Work:}

 
\paragraph{Single Target Domain Adaptation.} STDA is an unsupervised transfer learning task that focuses on adapting a model such that it can generalize well on an unlabeled target domain data while using a labeled source domain dataset. DL models for UDA seek to learn discriminant and domain-invariant representations from source and target data\cite{wangsurvey}. They are either based on either adversarial-\cite{GRL}, discrepancy-\cite{MMD_ICLR}, or reconstruction-based approaches\cite{DomainMapping1}. Taking advantage of adversarial training, several methods \cite{GRL, ADDA, CADA, ALDA} have been proposed using either gradient reversal\cite{GRL} or a combination of feature extractor and domain classifier to encourage domain confusion. Discrepancy-based approaches  \cite{MMD_ICLR, DA_Normalization_statistics} rely on measures between source and target distributions that can be minimized to generalize on the target domain. In \cite{MMD_ICLR}, authors minimize the Maximum Mean Discrepancy (MMD) between target and source features to find domain invariant features. On the other hand, \cite{DA_Normalization_statistics} assumes that task knowledge is already learned and the domain adaptation is done on the batch normalization layer to correct the domain shift. Lastly, another set of domain adaptation techniques focuses on the mapping of the source domain to target domain data or vice versa \cite{DomainMapping_Pixel_Level, ConGAN_DA_MAP}. These techniques are often based on the use of Generative Adversarial Network (GAN) in order to find a mapping between source and target.

\vspace{-3mm}
\paragraph{Knowledge Distillation (KD).} KD techniques allow for model compression by transferring knowledge from a teacher model, usually complex, to a smaller compact student model. The two main approaches of transferring knowledge between teacher and student models consist in minimizing the difference between logits \cite{HIntonKD, NoisyLabelKD}, and between features maps \cite{Yim_2017_CVPR, Fitnet, Overhaul}. Techniques from the first approach focus on measuring logits obtained from a temperature-based softmax and then minimize the distance between the logits of the teacher and the student \cite{HIntonKD}. More recently, techniques like \cite{Overhaul} minimize the distance between the intermediate feature maps of the teacher and student using a partial L2 distance. In contrast with other techniques, these features are obtained using a margin ReLU that accounts for negative values of the feature map. KD has been also recently employed in STDA \cite{KnowledgeAdaptation,IJCNN_KD_UDA}. For example, in \cite{KnowledgeAdaptation}, the authors use multiple teachers and employ a fusion scheme that sums the output of each teacher as distillation strategy. In a similar work \cite{IJCNN_KD_UDA}, STDA is performed during compression using knowledge distillation. While their approach is limited to a single-target scenario, we extend this approach to an MTDA setting by leveraging multi-teacher distillation into a single common student. 


\vspace{-3mm}
\paragraph{Multi Target Domain Adaptation.} MTDA is a set of domain adaptation techniques that improves upon the single target domain adaptation by adapting a single model to teacher target domains. Currently, MTDA still remains largely unexplored with many open research questions. The few existing MTDA approaches follow two main directions: MTDA either with target domain labels \cite{MTDA_Theoric} or without target domain labels \cite{compounddomainadaptation, BlendMTDA, dada}. The work in \cite{MTDA_Theoric} proposes an approach that can adapt a model to multiple target domains by maximizing the mutual information between domain labels and domain-specific features while minimizing the mutual information between the shared features. Recently, \cite{BlendMTDA} proposed to blend multiple target domains together and minimize the discrepancy between the source and the blended targets. Additionally, the authors employ an unsupervised meta-learner in combination with a meta target domain discriminator in order to blend the target domains. In \cite{compounddomainadaptation}, authors use a curriculum domain adaptations strategy combined with an augmentation of the representation based on features from source domain to handle multiple-target domains. While these methods achieve good performance, they fail to take advantage of existing STDA techniques, which have been extensively studied. Another important common point to existing methods is that they try to capture the representation of all the target domains using a common feature extractor directly from the data, which can degrade the final accuracy because of the limited capacity of the common model. In our paper, we overcome this issue by performing UDA separately on different models, and then distilling compressed knowledge to a common model. In addition, our experiments show that current mixed-target approaches still struggle with blending target domains in the feature space. We can gain more by preserving each domain specificity using STDA on different models.



\vspace{-2mm}
\section{Proposed Method}

\subsection{Domain Adaptation of Teachers:}

In this paper, the RevGrad \cite{GRL} technique is employed since it is the basis for many popular methods \cite{SegDA, FRCNN_DA}, although it can be easily replaced by other STDA techniques. Let us define the source domain as $S = \{x_s, y_s\}$ where $x_s$ is input pattern, and $y_s$ its corresponding label. The set of target domains is defined as $T = \{T_1, T_2, ... T_n\}$, each one defined as $T_i = \{x_t^i\}$. For each target domain $T_i$, we define a teacher model $\Phi_{i}$, and each of these teachers will be adapted to a corresponding target domain using the UDA technique proposed in \cite{GRL}. The domain adaptation of the teacher relies on a domain classifier, a gradient reversal layer (GRL), and the domain confusion loss:
\begin{equation}\label{eq:teacher_dc}
\footnotesize
	\begin{aligned}
	\mathcal{L}_{DC}(\phi_i, S, T_i) = \frac{1}{N_s + N_{ti}} \sum_{x \in S \cup T_i}\mathcal{L}_{CE}(D_i(\phi_i(x)), d_l)
	\end{aligned}
\end{equation}
where $\phi_i(x)$ is the output from the feature extractor of teacher network $\Phi_i$, before the fully connected layers, $D_i$ is the domain classifier for the corresponding teacher network, $d_l$ the domain label (source or target), $N_s$ is the number of samples in the source domain $S$, and $N_{ti}$ is the number of samples in the target domain $T_i$. 

\begin{figure}[htbp]
    \centering
    \includegraphics[width=\columnwidth]{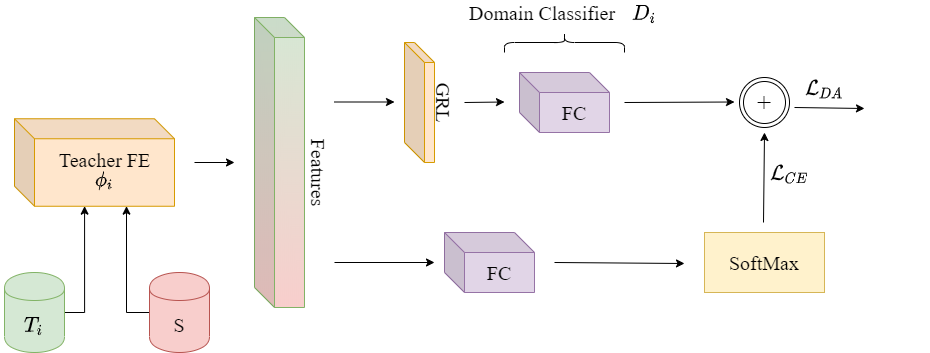}
    \caption{Illustration of GRL applied on a teacher model.}
    \label{fig:GRL_Teacher}
\end{figure}

The final domain adaptation loss is then defined as:
\begin{equation}\label{eq:teacher_da}
\footnotesize
	\begin{aligned}
	\mathcal{L}_{DA}(\Phi_i, S, T_i) = \frac{1}{N_s}\sum_{x_s, y_s \in S}\mathcal{L}_{CE}(\Phi_{i}(x_s),y_s) + \gamma \cdot \mathcal{L}_{DC}(\phi_i, T_i)
	\end{aligned}
\end{equation}
The first term (cross-entropy loss) allows the supervised training of the teacher model on the source domain that ensures the consistency of domain confusion. The second term is controlled by a hyper-parameter $\gamma$ that regulates the importance of the domain confusion loss which is maximized using a gradient reversal layer. Figure \ref{fig:GRL_Teacher} illustrates how GRL is applied for UDA.



\subsection{Teacher to Student Knowledge Distillation:}

\begin{figure}[htbp]
    \centering    \includegraphics[width=\columnwidth]{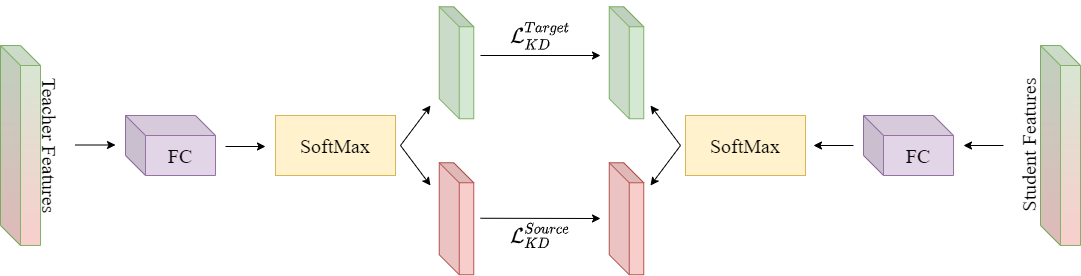}
    \caption{Illustration of proposed KD for domain adaptation.}
    \label{fig:KD}
\end{figure}

In this paper, we employ knowledge distillation based on logits as in \cite{HIntonKD}\footnotemark. The Figure \ref{fig:KD} illustrates the overall process of distillation on both target and source domains. Logits from a teacher/student model are fed to a temperature-based softmax function, in combination with a KL divergence loss on both the teacher and student outputs:
\begin{equation}\label{eq:source_kd}
	\footnotesize
	\begin{aligned}
	\mathcal{L}_{KD}^{Source}(\Phi_i, \Theta, S) = \frac{1}{N_s}\sum_{x_s, y_s \in S}\mathcal{L}_{KL}(\Phi_i(x_s, \tau), \Theta(x_s, 1)) \\ + \alpha \cdot  \mathcal{L}_{CE}(\Theta(x_s, 1), y_s)
	\end{aligned}
\end{equation}
\footnotetext{Note that our method can work with any other technique.}

where $\Theta$ represents our student model with $\tau$ the temperature hyper-parameter the softmax, and $\alpha$ the hyper-parameter to regulate the importance of the cross-entropy term. Even though the second term of Eq. \ref{eq:source_kd} may perform well with data from the source domain because it has labels, we add the domain confusion loss (Eq. \ref{eq:teacher_dc}) on the target domain to provide consistency during target distillation:
\begin{equation}\label{eq:target_kd}
	\footnotesize
	\begin{aligned}
	\mathcal{L}_{KD}^{Target}(\Phi_i, \Theta, T_i) = \frac{1}{N_{ti}}\sum_{x\in T_i}\mathcal{L}_{KL}(\Phi_i(x, \tau), \Theta(x, 1)) \\ + \alpha \cdot  \mathcal{L}_{DC}(\Theta, T_i)
	\end{aligned}
\end{equation}

\subsection{Multi-Teacher Multi-Target DA:}

For progressive UDA of teacher models and transfer of knowledge from teacher to the student model, we adapt an exponential growing rate to gradually transfer the importance of UDA to KD in a similar way to \cite{IJCNN_KD_UDA}. The growth rate is defined as:
\begin{equation}\label{eq:growth_rate}
\footnotesize
	\begin{aligned}
	g = \frac{\log(f/s)}{N_{e}}
	\end{aligned}
\end{equation}
where $s$ is the starting value, $f$ the final value, and $N_{e}$ the number of total epochs. This growth rate is used to calculate $\beta = s \cdot \exp \{g \cdot e\}$ with $e$ the current epoch in the overall loss function for optimization of one teacher:
\begin{equation}\label{eq:teacher_kd_grl_loss}
\footnotesize
	\begin{aligned}
	\mathcal{L}(\Phi_i, \Theta, S, T_i) = (1 - \beta)\mathcal{L}_{DA}(\Phi_i, T_i) + \\  \beta(\mathcal{L}_{KD}^{Source}(\Phi_i, \Theta, S)  + \mathcal{L}_{KD}^{Target}(\Phi_i, \Theta, T_i))
	\end{aligned}
\end{equation}

With $\beta$, the value that balances between the importance of the domain adaptation loss and the distillation loss. Our approach, MT-MTDA, instead of using deterministic fusion functions, such as average fusion, employs an alternative learning scheme for knowledge distillation from multiple teachers. This alternative scheme is performed by sequentially looping through each teacher at batch level (see Algorithm \ref{KD-UDA-Algo}).
\begin{algorithm}[htbp]
\footnotesize
\SetAlgoLined
\caption{Multi-Teacher Multi-Target Domain adaptation (MT-MTDA)}
\label{KD-UDA-Algo}
\SetKwInOut{Input}{input}
\SetKwInOut{Output}{output}
\SetKwInOut{Parameter}{parameter}
\Input{A source domain dataset $S$, a set of target dataset $T_0, T_1, ... T_n$}
\Output{A student model adapted to $n$ targets}

Initialize a set of teachers models $\Phi = \{\Phi_0, \Phi_1, ...\Phi_n\}$\\
Initialize a student model $\Theta$ \\

\For {$e\gets{1}$ \KwTo {$N_{e}$}} {
    \For {$x_s \in S$ and $X_t \in \{T_0,...T_n\}$} {
        Get the set of data of target domains $X_t$ \\
        \For {$x^{i}_{t} \in X_t$ and $\Phi_i \in \Phi$} {
            Optimize  $(1-\beta)\mathcal{L}_{DA}$ (\ref{eq:teacher_da}) for $\Phi_i$ using $x_s, x_t^i$ \\
            Optimize the loss of equation $\beta\mathcal{L}_{KD}^{Source}$ (\ref{eq:source_kd}) for $\Phi_i$ and $\Theta$ using $x_s$ \\
            Optimize the loss of equation $\beta\mathcal{L}_{KD}^{Target}$ (\ref{eq:target_kd}) for $\Phi_i$ and $\Theta$ using $x_t^i$ \\
        }
        Update $\beta = s \cdot \exp^{g \cdot e}$
    }
    Evaluate the model\\
}

\end{algorithm}

\begin{figure}[htbp]
    \centering
    \includegraphics[width=\columnwidth]{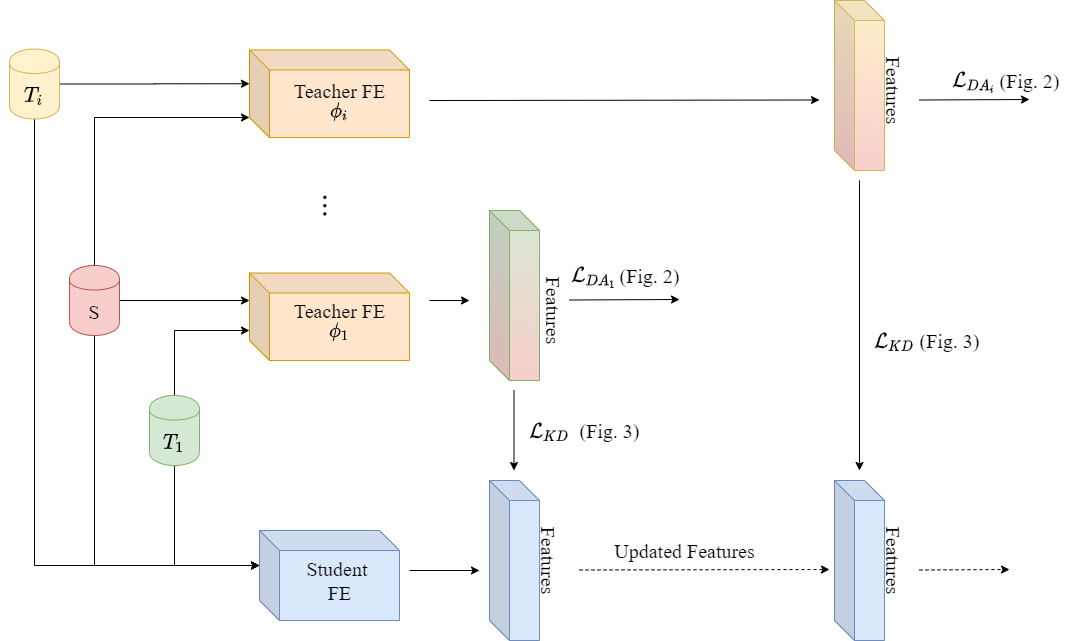}
    \caption{Illustration of the proposed learning technique.}
    \label{fig:MT-MTDA}
\end{figure}

Figure \ref{fig:MT-MTDA} illustrates the overall pipeline for our MT-MTDA approach. While all teachers share the same source dataset, the figure shows that they each teacher has its own target dataset with their own domain adaptation loss.

\vspace{-2mm}
\section{Experiments}

\subsection{Datasets:}

To the best of our knowledge, no specific dataset has been created for the for the MTDA task. For validation, we rely on datasets that are commonly used in other MTDA research \cite{BlendMTDA, MTDA_Theoric}, which are described below.

\vspace{-4mm}
\paragraph{1) Digits:} This dataset regroups a set of digits datasets: MNIST (\textbf{mt}), MNIST-M (\textbf{mm}), SVHN (\textbf{sv}), USPS (\textbf{up}) and Synthetic Digits (\textbf{sy}). Each one has each 10 classes that represent all the digits. For the evaluation on this dataset, we follow the same protocol as in \cite{BlendMTDA} for a fair comparison. 


\vspace{-4mm}
\paragraph{2) Office31:} It has 3 subsets -- Amazon (\textbf{A}), DSLR (\textbf{D}) and Webcam (\textbf{W}). These datasets all have 31 common classes. Images are taken respectively from the Amazon website, a DSLR camera and a webcam. We followed the standard evaluation protocol, a domain is chosen as a source, and the rest as targets. 




\vspace{-4mm}
\paragraph{3) OfficeHome:} This dataset contains 4 subsets: Art (\textbf{Ar}), Clip Art (\textbf{Cl}), Real World (\textbf{Rw}) and Product (\textbf{Pr}). It has a total of 15,500 images for 65 object categories that are usually found in office or home settings. We follow the same evaluation protocol of Office31.
\vspace{-4mm}
\paragraph{4) PACS:} While this dataset is often used for domain generalization, \cite{MTDA_Theoric} used it for MTDA. It contains 4 subsets: Art painting (\textbf{Ap}), Photo (\textbf{P}), Cartoon (\textbf{Cr}) and Sketch (\textbf{S}).

\vspace{-1mm}
\subsection{Implementation Details:}
In MT-MTDA, we use the same number of optimizers as teacher models, which are responsible for the UDA of each teacher. Additionally, we add another optimizer for the knowledge distillation of the student. MT-MTDA is compared to 1) a lower bound, which is only trained on source and tested on target, 2) the current state-of-the-art in MTDA with domain labels such as MTDA-ITA\cite{MTDA_Theoric} and 3) MTDA without domain labels such as AMEANS\cite{BlendMTDA}. Additionally, we compare to baseline methods such as RevGrad\cite{GRL} which is the basis of our MTDA method. We also use other baselines like DAN\cite{MMD_ICLR} or ADDA\cite{ADDA} in some cases for additional comparison, to show the advantages of MTDA algorithms. These baselines are domain adapted to directly to an ensemble of target domains similar to \cite{BlendMTDA}. For the Digits-five dataset, we employ a LeNet backbone with ResNet50 as teacher. As for the comparison on Office31 and OfficeHome, we use AlexNet backbone with ResNet50 as teacher models, and as for the comparison on the ResNet50 backbone, we use a ResNext101 as teachers. Our backbone CNNs follows the choices in \cite{BlendMTDA}. All these models start with pre-trained weights from ImageNet, except for LeNet. 

When comparing with AMEANS\cite{BlendMTDA} and DADA\cite{dada}, we add \textbf{MT-MTDA Mixed} -- our method when it employs mixed target domains without target domain labels. Specifically, we mix data from all the target domains, split them into subset of equal size without using domain labels, and then directly used them for UDA of respective teachers. For a fair comparison with AMEANS\cite{BlendMTDA}, we chose the same number of clusters, which corresponds to the number of target domains.

We selected the models' hyper-parameters based on their overall result in cross-validation in all the scenarios, instead of having a set of dedicated hyper-parameters for each scenario. Details of our hyper-parameters can be found in the Suppl. Material. We report the average classification accuracy obtained by all implemented models over 3 replications, from all the target domains. For other baselines, we report their best published result for fair comparison. Additional results (with OCDA\cite{compounddomainadaptation}, DADA\cite{dada}) and ablation studies (fusion methods, number of splits, etc.) are shown and analysed, along with a weighted average accuracy version of MT-MTDA in the Supplementary Material.

\vspace{-2mm}
\subsection{Results and Discussion:}


\paragraph{Comparison without domain labels}

\begin{table*}[htbp]
\centering
\caption{Accuracy of MT-MTDA and reference methods on the Digits-Five dataset.}
\label{tb:digits5_comp}
\resizebox{0.85\textwidth}{!}{
\begin{tabular}{|l|r|r|r|r|r|r|}
\hline
\textbf{Models}         & \textbf{mt $\xrightarrow{}$ mm, sv,} & \textbf{mm $\xrightarrow{}$ mt, sv,}   & \textbf{sv $\xrightarrow{}$ mt, mm,} & \textbf{sy $\xrightarrow{}$ mt, mm}    & \textbf{up $\xrightarrow{}$ mt, sv,} & \textbf{Avg} \\  
                        & \textbf{up, sy}                      & \textbf{up, sy}                      & \textbf{up, sy}                       & \textbf{up, sv}     & \textbf{mm, sy} &  \\ \hline \hline
Lower-bound: Superv. (source only)    & 36.6                            & 57.3                            & 67.1                           & 74.9                            & 36.9                           & 54.6    \\ \hline
ADDA\cite{ADDA}           & 52.5                            & 58.9                            & 46.4                           & 67.0                            & 34.8                           & 51.9    \\ \hline
DAN\cite{MMD_ICLR}            & 38.8                            & 53.5                            & 55.1                           & 65.8                            & 27.0                           & 48.0    \\ \hline
RevGrad\cite{GRL}         & 60.2                            & 66.0                            & 64.7                           & 69.2                            & 44.3                           & 60.9    \\ \hline
DADA\cite{dada} & 39.4 & 61.1 & \textbf{80.1} & \textbf{83.7} & 47.2 & 62.3 \\ \hline
AMEANS\cite{BlendMTDA}          & \textbf{61.2}                            & 66.9                            & 67.2                           & 73.3                            & 47.5                           & 63.2    \\ \hline 
MT-MTDA Mixed (ours) & 59.5 & \textbf{71.5} & 69.9 & 78.3 & 49.6 & \textbf{65.8} \\ \hline
MT-MTDA (ours) & 58.6                            & 71.0                            & 67.6                           & 75.6                            & \textbf{51.0}                           & 64.7    \\  \hline
Upper-bound: Superv. (targets) & 88.1 &	90.2 &	93.0 &	90.3 &	89.1 &	90.1 \\ \hline
\end{tabular}
}
\vspace{-4mm}
\end{table*}

Table \ref{tb:digits5_comp} shows the average classification accuracy of the MT-MTDA versus baseline and state-of-the-art methods on the Digits-Five dataset. We observe that our technique provides a higher level of accuracy, on average than the other approaches. In the first scenario, where our method performs poorly, further analysis on separate target domains (in Suppl. Material) indicates that our teacher models did not adapt well to the $mm$ and $sy$ datasets. This is mainly due to our selection of hyper-parameters, which was based on the all-scenario setting instead of individual cases. This explains why, in our first scenario, the result lags behind current baselines. It is possible, however, to overcome this problem by optimizing each scenario with a different set of hyper-parameters, including each teacher. Comparing the performance between the MT-MTDA Mixed (without domain labels) and MT-MTDA (with domain labels), the former improved slightly which might be due to the fact that the Mixed version is less susceptible to the sharing of hyper-parameters. Finally, an upper bound is included, i.e., trained and tested on target domain data, to show the gap between with a supervised model. 



\begin{table}[htbp]
\centering
\caption{Accuracy of MT-MTDA and reference methods on Alexnet and ResNet50 as backbone(student) on the Office31.}
\label{tb:Office31_comp}
\resizebox{\columnwidth}{!}{
\begin{tabular}{|l|r|r|r|r|}
	\hline
	\textbf{Models}          & \textbf{A $\xrightarrow{}$ D,W} & \textbf{D $\xrightarrow{}$ A,W} & \textbf{W $\xrightarrow{}$ A,D} & \textbf{Avg} \\ \hline \hline
		\multicolumn{5}{|c|}{\textbf{Teacher: ResNet50 --- Student: AlexNet}} \\ \hline
	Superv. (source only)       & 62.7                            & 73.3                            & 74.4                            & 70.1             \\ \hline
	DAN\cite{MMD_ICLR}                                         & 68.2                            & 71.4                            & 73.2                           & 70.9             \\ \hline
	RevGrad\cite{GRL}                                    & 74.1                            & 72.1                            & 73.4                            & 73.2             \\ \hline
	AMEANS\cite{BlendMTDA}                                     & 74.9                            & 74.9                            & 76.3                            & 75.4             \\ \hline
	MT-MTDA Mixed (Ours)                             & 80.3                   & \textbf{76.3}                   & \textbf{78.0}                   & \textbf{78.2}    \\ \hline 
	MT-MTDA (Ours)                             & \textbf{82.5}                   & 74.9                   & 77.6                  & \textbf{78.3}    \\ \hline \hline
	
	\multicolumn{5}{|c|}{\textbf{Teacher: ResNext101 --- Student: ResNet50}} \\ \hline
	Superv. (source only)     & 68.7                            & 79.6                            & 80.0                            & 76.1             \\ \hline
	DAN\cite{MMD_ICLR}                                        & 77.9                            & 75.0                            & 80.0                            & 77.6             \\ \hline
	RevGrad\cite{GRL}                                    & 79.0                            & 81.4                            & 82.3                            & 80.9             \\ \hline
	AMEANS\cite{BlendMTDA}                                     & \textbf{89.8}                            & \textbf{84.6}                            & \textbf{84.3}                            & \textbf{86.2}             \\ \hline
	MT-MTDA Mixed (Ours)                             & 85.5                           & 84.0                           & \textbf{84.4}                            & 84.6             \\ \hline
	MT-MTDA (Ours)                             & 87.9                            & 83.7                            & 84.0                            & 85.2             \\ \hline
	
\end{tabular}
}
\vspace{-4mm}
\end{table}


\begin{table*}[h!]
\centering
\caption{Accuracy of proposed and reference methods on OfficeHome dataset.}
\label{tb:OfficeHome_comp}
\resizebox{0.85\textwidth}{!}{
\begin{tabular}{|l|r|r|r|r|r|}
	\hline
	\textbf{Models}          & \textbf{Ar $\xrightarrow{}$ Cl, Pr, Rw} & \textbf{Cl $\xrightarrow{}$ Ar, Pr, Rw} & \textbf{Pr $\xrightarrow{}$ Ar, Cl, Rw} & \textbf{Rw $\xrightarrow{}$ Ar, Cl, Pr} & \textbf{Avg} \\ \hline \hline
	\multicolumn{6}{|c|}{\textbf{Teacher: ResNet50 --- Student: AlexNet}} \\ \hline
	Superv. (Source only)       & 33.4                                    & 35.3                                    & 30.6                                    & 37.9                                    & 34.3             \\ \hline
	DAN\cite{MMD_ICLR}                                       & 39.7                                    & 41.6                                    & 37.8                                    & 46.8                                    & 41.5             \\ \hline
	RevGrad\cite{GRL}                                    & 42.2                                    & 43.8                                    & 39.9                                    & 47.7                                    & 43.4             \\ \hline 
	AMEANS\cite{BlendMTDA}                                     & 44.6                                    & 45.6                                    & 41.4                                    & 49.3                                    & 45.2             \\ \hline
	MT-MTDA Mixed (Ours)                             & 48.6                         & 46.6                           & 41.1                           & 52.1                          & 47.1    \\ \hline 
	MT-MTDA (Ours)                             & \textbf{48.8}                           & \textbf{48.7}                           & \textbf{42.9}                           & \textbf{55.8}                           & \textbf{49.1}    \\ \hline \hline
	\multicolumn{6}{|c|}{\textbf{Teacher: ResNext101 --- Student: ResNet50}} \\ \hline
	Superv. (Source only)       & 47.6                                    & 41.8                                    & 43.4                                    & 51.7                                    & 46.1             \\ \hline
	DAN\cite{MMD_ICLR}                                        & 55.6                                    & 55.1                                    & 47.8                                    & 56.6                                    & 53.8             \\ \hline
	RevGrad\cite{GRL}                                    & 58.4                                    & 57.0                                    & 52.0                                    & 63.0                                    & 57.6             \\ \hline
	AMEANS\cite{BlendMTDA}                                     & 64.3                                    & 64.2                                    & 59.0                                      & 66.4                                    & 63.5             \\ \hline
	MT-MTDA Mixed (Ours)                             & \textbf{64.8}                         & 65.3                           & \textbf{60.5}                           & \textbf{67.5}                          & \textbf{64.5}    \\ \hline
	MT-MTDA (Ours)                             & 64.6                           & \textbf{66.4}                           & 59.2                           & 67.1                           & \textbf{64.3}    \\ \hline
\end{tabular}
}
\vspace{-4mm}
\end{table*}

Tables \ref{tb:Office31_comp} and \ref{tb:OfficeHome_comp} present the average classification accuracy of the MT-MTDA versus baseline and state-of-the-art methods on Office31 and OfficeHome data, respectively. In both cases, we observe that MT-MTDA, Mixed or with target domain labels, typically outperform the current state-of-the-art methods. With the AlexNet backbone, the improvements are significant, which can be explained by the advantage of using KD from multiple complex teachers, leading to a better generalization on a single target domain. We can observe that on Office31, AMEANS performs slightly better that MT-MTDA with the larger ResNet50 backbone. We believe that this is due to the limitations of domain adaptation on teacher models with MT-MTDA. We further discuss this point in the ablation study where we compare and discuss the performance of teacher and student (Sec. \ref{sec:teachers_vs_students}). Using our Mixed version on both these datasets, we often achieve similar results to the version with target domain labels.


\begin{table}[htbp]
\centering
\caption{Comparison with \cite{MTDA_Theoric} on PACS dataset.}
\label{tb:pacs_comp_mtda}
\resizebox{\columnwidth}{!}{
\begin{tabular}{|l|r|r|r|r|r|r|r|r|}
\hline
\textbf{LeNet}    & \multicolumn{4}{c|}{\textbf{P $\xrightarrow{}$ Ap, Cr, S}}                       & \multicolumn{4}{c|}{\textbf{Ap $\xrightarrow{}$ Cr, S, P}}                       \\ 
         & \textbf{P $\xrightarrow{}$ Ap} & \textbf{P $\xrightarrow{}$ Cr} & \textbf{P $\xrightarrow{}$ S} & \textbf{Avg}  & \textbf{Ap $\xrightarrow{}$ Cr} & \textbf{Ap $\xrightarrow{}$ S} & \textbf{Ap $\xrightarrow{}$ P} & \textbf{Avg}  \\ \hline \hline
ADDA \cite{ADDA}    & 24.3              & 20.1              & 22.4              & 22.3 & 17.8              & 18.9              & 32.8              & 23.2 \\ \hline
MTDA-ITA \cite{MTDA_Theoric} & \textbf{31.4}              & 23.0              & 28.2              & 27.6 & 27.0              & 28.9              & \textbf{35.7}              & 30.5 \\ \hline \hline
MT-MTDA (Ours)  & 24.6              & \textbf{32.2}              & \textbf{33.8}              & \textbf{30.2} & \textbf{46.6}              & \textbf{57.5}              & \textbf{35.6}              & \textbf{46.6} \\ \hline
\end{tabular}
}
\vspace{-4mm}
\end{table}

\vspace{-4mm}
\paragraph{Comparison with domain labels}
From Table \ref{tb:pacs_comp_mtda}, we observe that, in a comparison with another MTDA technique that uses domain labels\cite{MTDA_Theoric}, our method can have an improvement up to 15-16\%. Similar to other comparisons, the boost in our performance is due to the use of teacher models with higher capacity to generalize then distilled to the student.

\begin{figure}[htbp]
    \centering
    \includegraphics[width=\columnwidth]{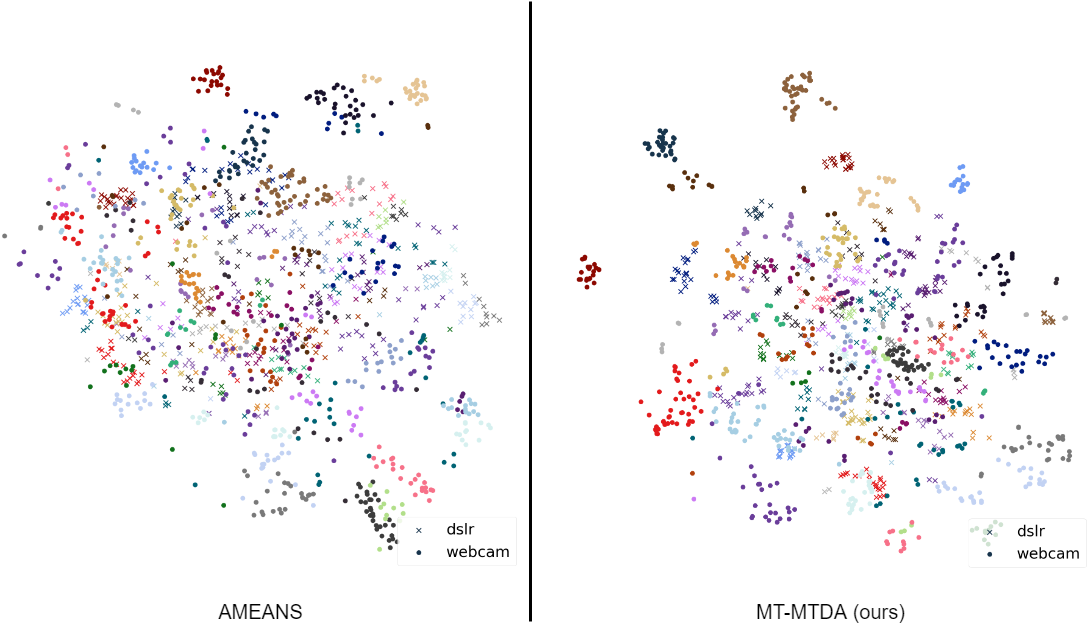}
    \caption{T-SNE visualization of Office31 data, where features are learned using MT-MTDA and AMEANS. Best viewed in color}
    \label{fig:TSNE_MTDA}
    \vspace{-3mm}
\end{figure}

Overall, our MT-MTDA technique outperforms both the baselines and state-of-the-art techniques. From Tables \ref{tb:digits5_comp}, \ref{tb:Office31_comp}, \ref{tb:OfficeHome_comp} and \ref{tb:pacs_comp_mtda}, we noticed that our model generally provides the highest accuracy on a compact backbone CNN, mainly because of the teacher's complexity and our knowledge distillation process. This is further confirmed by a comparison with the baseline RevGrad\cite{GRL} technique adapted directly on multi-target domains. Additionally, the improvements in accuracy that our methods brings decrease when the complexity gap between the teachers and student is smaller. In this case, the performance bottleneck is the teacher and the distillation algorithm. We further discuss this point in the ablation study, when comparing between student and teacher models. Finally, the difference between our versions is very small, meaning that our technique can perform well without domain labels and still preserve a high level of accuracy. This indicates that our teacher models can generalize well on multiple sub-mixtures of  target domains without reducing performance.

From Figure \ref{fig:TSNE_MTDA} \footnote{See a higher resolution image in Suppl. Material}, we observe that features learned with MT-MTDA better separates Office31 features, compared to other reference methods. Furthermore, MT-MTDA also separates class samples from different target domains in a better way than AMEANS. For comparison purposes, representations of other baselines are provided in the Suppl. Material. We noted that in current state-of-the-art method, the target domains do not blend well with each other since the feature extractor can still differentiate them quite well based on the t-SNE.

\subsection{Ablation Study}

\paragraph{Detailed Comparison of Each Target Domain.} For this experiment, we compare MT-MTDA in a setting where each target domain has a specific model. Our result on separate target domains is compared with RevGrad \cite{GRL}, but trained on a single target domain adaptation task. We also compared with the current best STDA algorithm, to our knowledge, in order to evaluate the effect of having a better STDA for the teacher model in our method.

\begin{table}[!htbp]
\centering
\caption{Accuracy of STDA methods for individual targets vs MTDA methods on Office31 dataset using AlexNet}
\label{tb:each_target_all_office31}
\resizebox{\columnwidth}{!}{
\begin{tabular}{|l|r|r|r|r|r|r|r|r|r|}
	\hline
	AlexNet & \multicolumn{3}{c|}{A $\xrightarrow{}$ D,W} & \multicolumn{3}{c|}{D $\xrightarrow{}$ A W} & \multicolumn{3}{c|}{W $\xrightarrow{}$ A D} \\ 
	             & A $\xrightarrow{}$ D & A $\xrightarrow{}$ W & Avg              & D $\xrightarrow{}$ A & D $\xrightarrow{}$ W & Avg  & W $\xrightarrow{}$ A & W $\xrightarrow{}$ D & Avg  \\ \hline \hline
	Cosine Distance      & 0.23                 & 0.21                 & 0.22             & 0.21                 & 0.24                 & 0.22 & 0.21                 & 0.24                 & 0.22 \\ \hline \hline
	RevGrad STDA & 72.3                 & 73.0                 & 72.6             & 53.4                 & 96.4                 & 74.9 & 51.2                 & 99.2                 & 75.2 \\ \hline
	DM-ADA\cite{DM_ADA}       & 77.5                 & 83.9                 & 80.7             & \textbf{64.6}                 & \textbf{99.8}                 & \textbf{82.2} & 64.0                 & \textbf{99.9}                 & \textbf{81.9} \\ \hline \hline
	AMEANS       & 74.7\footnotemark    & 74.6\footnote[2]     & 74.6\footnote[2] & -                    & -                    & 74.9    & -                    & -                    & 76.3    \\ \hline
	MT-MTDA Mixed      & 81.9                 & 78.7                & 80.3             & 56.2                 & 96.5                 & 76.3 & 56.3                 & 99.8                 & 78.0  \\ \hline
	MT-MTDA      & \textbf{83.1}                 & \textbf{81.9}                 & \textbf{82.5}             & 52.5                 & 97.3                 & 74.9 & 55.5                 & \textbf{100}                  & 77.6 \\ \hline
\end{tabular}
}
\vspace{-4mm}
\end{table}
\footnotetext{These results were obtained on the model provided in the authors's github, hence it is slightly different from the result reported in the original paper used in Table \ref{tb:Office31_comp}}

Table \ref{tb:each_target_all_office31} shows that while our algorithm does not perform as well as the state-of-the-art in STDA, it requires less computational power and memory since it only uses one model for all the target models instead of having a specific model for each target domain, i.e. two models in this case. This means that MTDA methods would typically scale better than current STDA methods since the number of models does not depend on the number of target models. Additionally, DM-ADA\cite{DM_ADA} shows that our algorithm can still be further improved since we can replace the STDA algorithm we are using on the teacher models (RevGrad\cite{GRL}) with almost any other STDAs, i.e. CDAN\cite{CADA}.  The table also shows the cosine distance in order to quantify the domain shift between source and target features for each UDA problem. Results show that the UDA problems in Office31 have a similar level of difficulty.

\vspace{-3mm}
\paragraph{Teachers vs Student Performance.}
\label{sec:teachers_vs_students}
We now compare the performances of our teachers with the student in order to explore the impact of knowledge distillation. For this experiment, we select the accuracy of each separate teacher on their respective target domain and compare it with the result of our student. This comparison was performed on the Office31 dataset\footnote{Additional results on OfficeHome can be found in the Supp. Material}. From Table \ref{tb:Teacher_vs_students}, the gap in accuracy is small and the student is almost at the same level as the teachers, except for the case of \textbf{A $\xrightarrow{}$ D,W}. This indicates that our student model has learned how to adapt to multi-target domains from each separate teacher without an explicit fusion scheme. The first scenario of \textbf{A $\xrightarrow{}$ D,W} shows a particular case where knowledge distillation helps improving domain adaptation. This behavior is also found when using ResNet50 as backbone architecture and seems to happen when the gap between the teachers and student is very small. This also suggests that knowledge from other teachers also help increasing accuracy. Additionally, from the ResNet50 backbone, we can see that the bottleneck can be found on teacher models and its domain adaptation since the student is stuck with a very similar accuracy as the teachers.

\begin{table}[!htbp]
\centering
\caption{Comparison with of teachers accuracy vs students}
\label{tb:Teacher_vs_students}
\resizebox{\columnwidth}{!}{
\begin{tabular}{|l|l|l|l|l|l|}
\hline
               & \textbf{Models}     & \textbf{A $\xrightarrow{}$ D,W} & \textbf{D $\xrightarrow{}$ A,W} & \textbf{W $\xrightarrow{}$ A,D} & \textbf{Average} \\ \hline \hline
Student & AlexNet    & \textbf{82.5}                & 74.9                & 77.6                & 78.3    \\ \hline
Teachers       & ResNet50   & 77.6                & \textbf{79.9}                & \textbf{80.0}                & \textbf{79.2}    \\ \hline \hline
Student & ResNet50   & \textbf{87.9}                & \textbf{83.7}                & 84.0                & \textbf{85.2}    \\ \hline
Teachers       & ResNext101 & 85.9                & 83.6                & \textbf{84.3}                & 84.6    \\ \hline
\end{tabular}
}
\vspace{-4mm}
\end{table}

\begin{table*}[!h]
\centering
\caption{Comparison of student's accuracy on separate domains with gradual increase in domains. The order in the target domains indicates the order in which they were integrated into the training.}
\label{tb:gradual_targets}
\resizebox{0.7\textwidth}{!}{
\begin{tabular}{|l|r|r|r|r|r|r|r|}
\hline
\textbf{AlexNet}           & \textbf{Ar $\xrightarrow{}$ Cl} & \textbf{Ar $\xrightarrow{}$ Pr} & \textbf{Ar $\xrightarrow{}$ Rw} & \textbf{Ar $\xrightarrow{}$ Pr, Cl}   & \textbf{Ar $\xrightarrow{}$ Pr, Rw} & \textbf{Ar $\xrightarrow{}$ Rw, Cl} & \textbf{Ar $\xrightarrow{}$ Cl, Pr, Rw} \\ \hline \hline
Acc on Cl & \textbf{34.0}                &                     &                     & 33.3                 &                         & 33.0                      & \textbf{34.1}                        \\ \hline
Acc on Pr &                     & \textbf{55.3}                &                     & 50.1                   & 50.0                      &                         & 52.6                        \\ \hline
Acc on Rw &                     &                     & 59.0                  &                       & 57.9                    & 57.7                    & \textbf{59.7}                        \\ \hline
\end{tabular}
\vspace{-4mm}
}

\end{table*}

\begin{table*}[htbp]
\centering
\caption{Accuracy of each target domain and standard deviation between these accuracies}
\label{tb:order}
\resizebox{0.7\textwidth}{!}{
\begin{tabular}{|l|r|r|r|r|r|r|c|}
\hline
\textbf{AlexNet}    & \textbf{Ar $\xrightarrow{}$ Cl,Pr,Rw} & \textbf{Ar $\xrightarrow{}$ Cl,Rw,Pr} & \textbf{Ar $\xrightarrow{}$ Pr,Cl,Rw} & \textbf{Ar $\xrightarrow{}$ Pr,Rw,Cl} & \textbf{Ar $\xrightarrow{}$ Rw,Cl,Pr} & \textbf{Ar $\xrightarrow{}$ Rw,Pr,Cl} & \textbf{STDev}         \\ \hline \hline
Acc on Cl & 34.1                      & 33.7                      & 34.0                          & \textbf{34.7}                      & 33.6                          & \textbf{34.6}                      & 0.4        \\ \hline
Acc on Pr & 52.6                      & 52.5                      & \textbf{53.1}                          & 52.5                      & 52.5                          & \textbf{53.1}                      & 0.3        \\ \hline
Acc on Rw & 59.7                      & \textbf{60.6}                      & 59.8                          & 59.5                      & 60.4                          & 59.5                      & 0.4        \\ \hline \hline
Average Acc   & 48.8                      & 48.9                      & 48.9                          & 48.9                      & 48.8                          & \textbf{49.1}                      & 0.1 | 0.3 \\ \hline
\end{tabular}
}
\vspace{-3mm}
\end{table*}

\vspace{-3mm}
\paragraph{Consistency on Target Knowledge Distillation.}
In this section, we evaluate the impact of the consistency loss on the target knowledge distillation \ref{eq:target_kd}. To this end, we removed the second term of the target knowledge distillation, Eq. \ref{eq:target_kd}, completely and we run our algorithm with the same settings as before on the scenario with an AlexNet as backbone on Office31 dataset. From the Table \ref{tb:Office31_no_cst}, it seems that having a consistency term on the target distillation loss only brings a small boost in performance. This aligns with our main results since the hyper-parameter that controls this consistency term is set to a small value.

\begin{table}[!h]
\centering
\caption{Accuracy of proposed method with target distillation consistency vs without}
\resizebox{\columnwidth}{!}{
\label{tb:Office31_no_cst}
\begin{tabular}{|l|r|r|r|r|}
\hline
\textbf{Models} & \textbf{A $\xrightarrow{}$ D,W} & \textbf{D $\xrightarrow{}$ A,W} & \textbf{W  $\xrightarrow{}$ A,D} & \textbf{Average}     \\ \hline \hline
MT-MTDA without CST & 82.1                & 73.8                & 74.3                & 76.7 \\ \hline
MT-MTDA with CST    & \textbf{82.5}                & \textbf{74.9}                & \textbf{77.6}                & \textbf{78.3} \\ \hline
\end{tabular}
}
\vspace{-3mm}
\end{table}

\vspace{-3mm}
\paragraph{Single Teacher vs Multiple Teachers.} We compare the scenario of having multiple teachers, each in a different target domain versus a scenario with one teacher adapting on a mixed target domain similar to \cite{IJCNN_KD_UDA}. In this setting, we merge all the target domains into a single target domain, where a single teacher is then assigned. We run this study on the Office31 dataset with the same hyper-parameters as in the main experiment. From the results of Table \ref{tb:Office31_mixed}, we can observe that the accuracy of a mixed target domain using similar algorithm to \cite{IJCNN_KD_UDA} is significantly lower than the results with a multi-teacher approach. This suggests that even with a complex teacher network, a good generalization on a mixed target domain is hard to achieve and a multi-teacher scenario is preferable.

\begin{table}[h!]
\centering
\caption{Accuracy of proposed method using a single teacher vs multiple teachers}
\label{tb:Office31_mixed}
\resizebox{\columnwidth}{!}{
\begin{tabular}{|l|r|r|r|r|}
	\hline
	\textbf{Models}      & \textbf{A $\xrightarrow{}$ D,W} & \textbf{D $\xrightarrow{}$ A,W} & \textbf{W  $\xrightarrow{}$ A,D} & \textbf{Average} \\ \hline \hline
	Single Teacher MTDA & 75.3                            & 64.0                            & 67.0                             & 68.8             \\ \hline
	MT-MTDA              & \textbf{82.5}                & \textbf{74.9}                & \textbf{77.6}                & \textbf{78.3}    \\ \hline
\end{tabular}
}
\vspace{-4mm}
\end{table}

\vspace{-3mm}
\paragraph{Impact of Number of Target Domains.}
We now investigate the impact of increasing the number of domains on the student model. This experiment starts with a STDA setting of our algorithm and slowly increases the number of domains until reaching the maximum. We decided to do this experiment on the scenario with \textbf{Ar} dataset as source in OfficeHome dataset since it has more than two target domains and the dataset is bigger than Digits-Five. From Table \ref{tb:gradual_targets}, we can see that while the performance degrades on the target domain \textbf{Pr}, we notice a slight increase in accuracy of the other cases. This means that, with our method, training multiple target domains together can boost the performance of some separate target domains. The decrease of performance in the case of \textbf{Pr} also indicates that there might be a saturation in learning capacity. In this case, we can say that the target domains \textbf{Cl} and \textbf{Rw} improved at the expense of \textbf{Pr}. 


\vspace{-4.5mm}
\paragraph{Order of Target Domains.} We evaluate whether the order of target domains impacts the performance of the final model. Similarly to the previous experiment, we used the scenario with \textbf{Ar} as the source domain in OfficeHome dataset since there are more than 2 target domains. Table \ref{tb:order} reports the results of individual target domains when their orders are different. These results indicate that even though the order of the domains leads to different average results, the difference between the configurations is marginal, of nearly $0.3\%$, with a standard deviation equal to $0.1$. These results indicate that the order of target domains has little impact, if any, on the final result of the trained models.


\vspace{-4mm}
\section{Conclusion}

In this paper, an avenue is unexplored for MTDA, relying on multiple teachers in order to distill knowledge from multiple target domains into a single student. Results from our experiment show that our method outperforms the current state-of-the-art, especially when using compact models, which can facilitate the use in numerous real-time applications. From our experiment, we identify several bottlenecks that can impede generalization of a compact model to multiple domains: 1) the STDA algorithm determines the accuracy of teacher models and 2) the transfer of target domain knowledge which needs to be improved when the student model is compact. Since STDA is a popular research area, our future work will focus on how to transfer target domain knowledge.

\vspace{-3mm}
\section*{Acknowledgment}
This research was supported in part by the NSERC, Compute Canada, and MITACS.

\newpage
{\small
\bibliography{sample}
}

\newpage
\appendix
\clearpage
{ \centering \Large \textbf{Supplementary Material}}

\section{Experimental Methodology}

\subsection{Hyper-parameters}

\begin{table*}[hbp]
\large
\centering
\caption{Hyper-parameters for our algorithms for each backbone and dataset}
\label{tb:hyper_parameters}
\resizebox{0.95\textwidth}{!}{
\begin{tabular}{|l|r|r|r|r|r|}
	\hline 
	\textbf{Hyper parameters} & \textbf{Digits-Five LeNet} & \textbf{Office31 Alexnet} & \textbf{OfficeHome Alexnet} & \textbf{Office31 ResNet50} & \textbf{OfficeHome Resnet50} \\ \hline \hline
	$N_{e}$                    & 100                        & 100                       & 200                         & 100                        & 200                          \\ \hline
	batch size                & 64                         & 16                        & 8                           & 16                         & 8                            \\ \hline
	$\tau$                    & 20                         & 20                        & 20                          & 20                         & 20                           \\ \hline
	$\alpha$                  & 0.5                        & 0.3                       & 0.5                         & 0.3                        & 0.5                          \\ \hline
	$s$                       & 0.1                        & 0.1                       & 0.1                         & 0.1                        & 0.1                          \\ \hline
	$f$                       & 0.8                        & 0.8                       & 0.5                         & 0.8                        & 0.5                          \\ \hline
	$\gamma$                  & 0.5                        & 0.5                       & 0.5                         & 0.5                        & 0.5                          \\ \hline
	UDA Learning Rate         & 0.0005                     & 0.001                     & 0.0001                      & 0.001                      & 0.0001                       \\ \hline
	KD Learning Rate          & 0.0005                     & 0.01                      & 0.001                       & 0.01                       & 0.001                        \\ \hline
	weight decay             & 0.0005                     & 0.0005                    & 0.0005                      & 0.0005                     & 0.0005                       \\ \hline
\end{tabular}
}
\end{table*}

From Table \ref{tb:hyper_parameters}, you can find all the hyper-parameters that was used for different datasets and backbones. We selected these hyper-parameters based on a standard cross-validation process. These hyper-parameters are selected based on the overall result in all the scenarios instead of each scenario.

\subsection{Evaluation metrics}

In the paper of \cite{BlendMTDA}, the authors first proposed an accuracy metrics that take in account the different sizes of each target domain in order to have a balanced accuracy score at the end. This accuracy is defined as:
\begin{equation}\label{eq:weighted_acc}
	\begin{aligned}
    Acc = \sum_{i=0}^{n}w_iAcc_i
	\end{aligned}
\end{equation}

With $w_i$ calculated as $w_i = \frac{N_i}{\sum_{j=0}^{n}N_j}$. The problem with this accuracy is that it can hide the poor performance of a target domain that is small. The authors from the same paper proposed to use another accuracy which is the same one we used in our main paper where the same weight is used for each target domain. This is also referred as the equal-weight classification accuracy in the paper of \cite{BlendMTDA}. This is accuracy is calculated as:
\begin{equation}\label{eq:ew_acc}
	\begin{aligned}
    Acc^{EQ} = \frac{1}{n}\sum_{i=0}^{n}Acc_i
	\end{aligned}
\end{equation}

Additionally, we also give our result based on the Equation \ref{eq:weighted_acc} and on each target domain in order to highlight where our algorithm can fail.

\section{Results and Discussion}
\subsection{Digits-Four}

In this comparison, we compare our results with both MTDA-ITA\cite{MTDA_Theoric} and \cite{compounddomainadaptation} on the same scenario as in \cite{compounddomainadaptation} on Digits. For a fair comparison with \cite{compounddomainadaptation}, which is an open domain adaptation algorithm, we only takes in account the compound domains results of OCDA (close domain adaptation) since they are part of the closed domain of OCDA which is similar to our setting.

\begin{table*}[htbp]
\large
\centering
\caption{Accuracy of proposed and reference methods on Digits-Four dataset}
\label{tb:digits4_ocda}
\resizebox{0.7\textwidth}{!}{
\begin{tabular}{|l|l|l|l|l|}
\hline
& \multicolumn{4}{|c|}{\textbf{Source $\rightarrow$ Targets}}  \\ 
\textbf{LeNet}    & \multicolumn{4}{c|}{\textbf{sv $\xrightarrow{}$ mt, mm, up}}                       \\ 
         & \textbf{sv $\xrightarrow{}$ mt} & \textbf{sv $\xrightarrow{}$ mm} & \textbf{sv $\xrightarrow{}$ up} & \textbf{Average}  \\ \hline \hline
ADDA \cite{ADDA}     & 80.1              & 56.8              & 64.8              & 67.2  \\ \hline
MTDA-ITA \cite{MTDA_Theoric}     & 84.6              & 65.3              & 70.0              & 73.3  \\ \hline
AMEANS \cite{BlendMTDA}     & 85.2              & 65.7              & 74.3              & 75.1  \\ \hline
OCDA \cite{compounddomainadaptation}     & 90.9              & 65.7              & 83.4              & 80.0  \\ \hline \hline
MD-MTDA Mixed (Ours)  & 87.5              & 65.4              & 84.7              & 79.2 \\ \hline
MD-MTDA (Ours)  &   86.9            & 65.2              & 84.3              &78.8 \\ \hline
MD-MTDA CDAN Mixed (Ours)  & \textbf{92.8}              & 67.3              & 85.1              & 81.7 \\ \hline
MD-MTDA CDAN (Ours)  & 92.0              & \textbf{71.1}              & \textbf{88.9}              &\textbf{84.0} \\ \hline

\end{tabular}
}
\end{table*}

From Table\ref{tb:digits4_ocda}, we observe that, while our base techniques (with RevGrad) perform slightly worse than OCDA, it still performs better than most of the other techniques. As for our version that uses CDAN\cite{CADA}, our technique perform even better than state-of-the-art technique \cite{compounddomainadaptation} by around 1\%. 

\subsection{Comparison with DADA\cite{dada} on Office-Caltech 10}

\begin{table*}[htbp]
\large
\centering
\caption{Accuracy of MT-MTDA and reference methods on Alexnet and Resnet101 as backbone(student) on the Office-Caltech}
\label{tb:Officecaltech_comp}
\resizebox{0.8\textwidth}{!}{
\begin{tabular}{|l|r|r|r|r|r|}
	\hline
	\textbf{Models}          & \textbf{A $\xrightarrow{}$ C,D,W}  & \textbf{C $\xrightarrow{}$ A,D,W} & \textbf{D $\xrightarrow{}$ A,C,W} & \textbf{W $\xrightarrow{}$ A,C,D} & \textbf{Average} \\ \hline \hline
		\multicolumn{6}{|c|}{\textbf{Teacher: ResNet50 --- Student: AlexNet}} \\ \hline
	Source only             & 83.1        & 88.9      & 86.7      & 82.2          & 85.2         \\ \hline
	RevGrad\cite{GRL}       & 85.9        & 90.5      & 88.6      & 90.4          & 88.9         \\ \hline
	DADA\cite{dada}         & 86.3        & 91.7      & 89.9      & 91.3          & 89.8         \\ \hline
	MT-MTDA Mixed (Ours)    & 92.8        & 93.4      & 89.2      & 90.8          & 91.6         \\ \hline 
	MT-MTDA (Ours)          & \textbf{93.3}        & \textbf{93.9}      & \textbf{90.1}      & \textbf{91.2}          & \textbf{92.1}         \\ \hline \hline

 	\multicolumn{6}{|c|}{\textbf{Teacher: ResNext101 --- Student: ResNet101}} \\ \hline
 	Source only             & 90.5        & 94.3      & 88.7      & 82.5          & 89.0         \\ \hline
 	RevGrad\cite{GRL}       & 91.5        & 94.3      & 90.5      & 86.3          & 90.6         \\ \hline
 	DADA\cite{dada}         & 92.0        & 95.1      & 91.3      & 93.1          & 92.9         \\ \hline
 	MT-MTDA Mixed (Ours)    & 94.9        & 97.9      & 94.7      & 95.3          & 95.7         \\ \hline 
 	MT-MTDA (Ours)          & \textbf{96.1}        & \textbf{98.1}      & \textbf{96.3}      & \textbf{96.4}          & \textbf{96.7}         \\ \hline
\end{tabular}
}
\end{table*}

In this experiment, we provide an additional comparison with DADA\cite{dada} on the Office-Caltech10 dataset with AlexNet and ResNet101 as backbones similar to DADA\cite{dada}. For this, we use similar hyper-parameters as the Office31 dataset. From Table \ref{tb:Officecaltech_comp}, we can see that both version of our techniques perform better than DADA\cite{dada}. Similar to previous comparison, we think that our technique achieves state-of-the-art performance by taking advantages of higher generalization capacity of teacher models and transfer it to a common student.

\subsection{Further analysis on Digits-Five}
As indicated in our results for Table \ref{tb:digits5_comp}, we analyzed the accuracy of each target domain in order to show where's our drop in accuracy.

\begin{table*}[htbp]
\centering
\caption{Accuracy of each target domain on Digits-Five dataset with LeNet as Backbone.}
\label{tb:Digits5_separate}
\resizebox{1.0\textwidth}{!}{
\begin{tabular}{|l|r|r|r|r|r|}
\hline
\textbf{Lenet}             & \textbf{mt $\xrightarrow{}$ mm, sv, up, sy} & \textbf{mm $\xrightarrow{}$ mt, sv, up, sy} & \textbf{sv $\xrightarrow{}$ mt, mm, up, sy} & \textbf{sy $\xrightarrow{}$ mt, mm, up, sv} & \textbf{up $\xrightarrow{}$ mt, sv, mm, sy} \\ \hline \hline
Student Acc on mt & -                                & 96.3                            & 69.1                           & 85.8                            & 87.0                             \\ \hline
Student Acc on mm & 46.6                            & -                                & 48.1                           & 55.5                            & 40.3                           \\ \hline
Student Acc on sv & 53.8                            & 43.9                            & -                               & 75.3                            & 30.7                           \\ \hline
Student Acc on sy & 57.7                            & 82.9                            & 83.7                           & -                                & 48.4                           \\ \hline
Student Acc on up & 77.3                            & 61.1                            & 69.4                           & 85.8                            & -                               \\ \hline \hline
Average           & 58.85                           & 71.05                           & 67.575                         & 75.6                            & 51.6                           \\ \hline
\end{tabular}
}
\end{table*}

From Table \ref{tb:Digits5_separate}, we can see that the drop in performance in the scenario previously noted in the main paper is due to the decline of the domain adaptation on both \textbf{mt $\xrightarrow{}$ mm} and \textbf{mt $\xrightarrow{}$ sy}. Further analysis of these two domain adaptation shows that our common hyper-parameters do no work well for these two cases since we can get better performance using other hyper-parameters. However, these parameters would yield lower performance on other scenarios therefore we choose to remain on the same hyper-parameters as before. This indicates that in order to have better performance, it is best to have a different hyper-parameters set for each scenario and even each teacher.

\subsection{Number of splits for mixed target domains}
In this experiment, we evaluate the impact of the number of splits with our technique on the Office31 dataset with AlexNet as backbone. We gradually increase the number of splits starting from two splits, since one splits is the same as the scenario of single-teacher with a mixture of targets.

\begin{table*}[!htbp]
\centering
\caption{Accuracy of MT-MTDA Mixed on different number of splits of sub-targets}
\label{tb:num_of_splits}
\resizebox{0.65\textwidth}{!}{
\begin{tabular}{|l|r|r|r|r|}
\hline
\textbf{AlexNet}          & \textbf{A $\xrightarrow{}$ D,W} & \textbf{D $\xrightarrow{}$ A,W} & \textbf{W $\xrightarrow{}$ A,D} & Average \\ \hline \hline
MT-MTDA Mixed 2 & 80.3                                     & 76.3                                     & 78.0                                     & 78.2                         \\ \hline
MT-MTDA Mixed 3 & 81.2                                     & 76.2                                     & 78.2                                     & 78.5                         \\ \hline
MT-MTDA Mixed 4 & \textbf{82.1}                                     & 76.6                                     & \textbf{78.8}                                     & \textbf{79.2}                         \\ \hline
MT-MTDA Mixed 10 & 81.3 & \textbf{76.9} & 78.5 & 78.9 \\ \hline
\end{tabular}
}
\end{table*}

From Table \ref{tb:num_of_splits}, we noticed that an increase in the number of splits and the number of teachers can result in a slight increase in the overall accuracy. The results also show that our teacher model is capable of adapting to randomly split sub-mixture of target domains and then transfer the knowledge to a single student model, independently from the number of splits. In addition, these results also show the robustness of our algorithm since the sub-mixture target domains are always obtained randomly.

\subsection{MT-MTDA using another STDA technique} 
In this experiment, we evaluate our algorithm with another domain adaptation technique, namely \cite{CADA}, in order to show that our algorithm is agnostic w.r.t domain adaptation technique. We will also do the same for distillation, where we use \cite{Overhaul} distillation.

\begin{table*}[htbp]
\centering
\caption{Accuracy of MT-MTDA with RevGrad vs CDAN}
\label{tb:Office31_comp_cdan}
\resizebox{0.7\textwidth}{!}{
\begin{tabular}{|l|r|r|r|r|}
	\hline
	\textbf{Models}          & \textbf{A $\xrightarrow{}$ D,W} & \textbf{D $\xrightarrow{}$ A,W} & \textbf{W $\xrightarrow{}$ A,D} & \textbf{Average} \\ \hline \hline
		\multicolumn{5}{|c|}{\textbf{Teacher: ResNet50 --- Student: AlexNet}} \\ \hline
	MT-MTDA Mixed (Ours)                             & 80.3                   & 76.3                   & \textbf{78.0}                   & 78.2    \\ \hline 
	MT-MTDA (Ours)                             & 82.5                   & 74.9                   & 77.6                  & 78.3    \\ \hline \hline
	
	MT-MTDA CDAN Mixed (Ours)                             & 84.3                   & 75.4                   & 77.8                   & 79.2    \\ \hline
	MT-MTDA CDAN (Ours)                             & \textbf{84.5}                   & \textbf{76.9}                   & \textbf{78.0}                  & \textbf{79.8}    \\ \hline 
\end{tabular}
}
\end{table*}


From Table \ref{tb:Office31_comp_cdan}, we noticed that the version with CDAN\cite{CADA}, while it does not present the same performance gap as in the STDA case as shown in \cite{CADA}, it still performs better than the version with RevGrad of around $1\%$. Taking in account result from \ref{tb:digits4_ocda}, we can see that our student model is reaching its limit in term of generalization across multiple domains for the Office31 dataset. Lastly, results in both of these tables also show the improvement our algorithm can achieve when employing state-of-the-art UDA technique.

\subsection{Comparison with Other Fusion Methods.} To demonstrate the benefits of the proposed feature fusion strategy, we compare our alternative fusion scheme with other baselines fusion methods, e.g., the sum or the mean of the output. The hyper-parameters for all the cases remain the same to those of the main experiment, with the only difference being the output of all the teachers is summed/averaged and then distill to the student. Table \ref{tb:Office31_fusion} shows that the proposed alternative distillation works better than either fusion by sum or average. This means that the proposed alternative scheme transfers learned knowledge better than the baseline methods in the particular case of MTDA. In addition, this shows that the student does not need an explicit fusion scheme in order to learn target domain knowledge from multiple teachers.

\begin{table*}[ht]
\centering
\caption{Accuracy of proposed method with different fusions}
\label{tb:Office31_fusion}
\resizebox{0.65\textwidth}{!}{
\begin{tabular}{|l|r|r|r|r|}
\hline
\textbf{Models} & \textbf{A $\xrightarrow{}$ D,W} & \textbf{D $\xrightarrow{}$ A,W} & \textbf{W  $\xrightarrow{}$ A,D} & \textbf{Average}     \\ \hline \hline
MT-MTDA Mean & 75.1 & 65.4 & 67.1 & 69.2 \\ \hline
MT-MTDA Sum  & 78.3 & 66.9 & 69.8 & 71.6 \\ \hline
MT-MTDA   & \textbf{82.5}                & \textbf{74.9}                & \textbf{77.6}                & \textbf{78.3} \\ \hline
\end{tabular}
}
\end{table*}

\subsection{Weighted Accuracy}
In this section, we present our average accuracy using Equation \ref{eq:weighted_acc}. We compare with the weighted accuracy reported in the paper of \cite{BlendMTDA} for a fair comparison.


\begin{table*}[htbp]
\centering
\caption{Weighted accuracy of proposed and baseline methods on Digits-Five dataset with AlexNet as Backbone.}
\label{tb:Digits5_comp_alexnet_weighted}
\resizebox{1.0\textwidth}{!}{
\begin{tabular}{|l|r|r|r|r|r|r|}
\hline
\textbf{Models}      & \textbf{mt $\xrightarrow{}$ mm, sv, up, sy} & \textbf{mm $\xrightarrow{}$ mt, sv, up, sy} & \textbf{sv $\xrightarrow{}$ mt, mm, up, sy} & \textbf{sy $\xrightarrow{}$ mt, mm, up, sv} & \textbf{up $\xrightarrow{}$ mt, sv, mm, sy} & \textbf{Average} \\ \hline \hline
Source only & 26.9                            & 56.0                            & 67.2                           & 73.8                            & 36.9                           & 52.2    \\ \hline
ADDA        & 43.7                            & 55.9                            & 40.4                           & 66.1                            & 34.8                           & 48.2    \\ \hline
DAN         & 31.3                            & 53.1                            & 48.7                           & 63.3                            & 27.0                           & 44.7    \\ \hline
RevGrad     & 52.4                            & 64.0                            & 65.3                           & 66.6                            & 44.3                           & 58.5    \\ \hline
AMEANS      & 56.2                            & 65.2                            & 67.3                           & 71.3                            & 47.5                           & 61.5    \\ \hline \hline
MT-MTDA Mixed (ours)        & 51.6                            & 69.2                           & \textbf{79.7} & \textbf{76.0}                          & 61.5                           & \textbf{67.6}                           \\ \hline
MT-MTDA (ours)        & \textbf{54.3}                            & \textbf{73.4}                            & 67.1                           & 73.1                            & \textbf{64.0}                           & 66.4    \\ \hline
\end{tabular}
}
\end{table*}

\begin{table*}[htbp]
\centering
\caption{Weighted accuracy of proposed and baseline methods on Office31 dataset.}
\label{tb:office31_comp_weighted}
\resizebox{0.7\textwidth}{!}{
\begin{tabular}{|l|r|r|r|r|r|}
\hline
\textbf{Models}      & \textbf{A $\xrightarrow{}$ D,W} & \textbf{D $\xrightarrow{}$ A,W} & \textbf{W $\xrightarrow{}$ A,D} & \textbf{Average} \\ \hline \hline
		\multicolumn{5}{|c|}{\textbf{Teacher: ResNet50 --- Student: AlexNet}} \\ \hline
Source only   & 62.4                & 60.8                & 57.2                & 60.1    \\ \hline
DAN\cite{MMD_ICLR}                                    & 68.2                & 58.7                & 55.6                & 60.8    \\ \hline
RevGrad\cite{GRL}                                & 74.1                & 58.6                & 55.0                & 62.6    \\ \hline
AMEANS\cite{BlendMTDA}                                 & 74.5                & 62.8                & 59.7                & 65.7    \\ \hline
MT-MTDA Mixed (ours)                                & 79.9                & \textbf{65.1}                & \textbf{62.8}                & \textbf{69.3}    \\ \hline
MT-MTDA (ours)                                & \textbf{82.4}                & 62.4                & 61.9                & 68.9    \\ \hline \hline
\multicolumn{5}{|c|}{\textbf{Teacher: ResNext101 --- Student: ResNet50}} \\ \hline
Source only  & 68.6                & 70.0                & 66.5                & 68.4    \\ \hline
DAN\cite{MMD_ICLR}                                    & 78.0                & 64.4                & 66.7                & 69.7    \\ \hline
RevGrad\cite{GRL}                                & 78.2                & 72.2                & 69.8                & 73.4    \\ \hline
AMEANS\cite{BlendMTDA}                                 & \textbf{90.1}                & \textbf{77.0}                & 73.4                & \textbf{80.2}    \\ \hline 
MT-MTDA Mixed (ours)                                & 85.2               & 75.8                & \textbf{73.6}               & 78.2    \\ \hline
MT-MTDA (ours)                                & 87.8                & 75.4                & 72.8                & 78.7    \\ \hline
\end{tabular}
}
\end{table*}

\begin{table*}[htbp]
\centering
\caption{Weighted accuracy of proposed and baseline methods on OfficeHome dataset.}
\label{tb:officehome_comp_weighted}
\resizebox{.8\textwidth}{!}{
\begin{tabular}{|l|r|r|r|r|r|}
\hline
Models                         & Ar -\textgreater Cl, Pr, Rw & Cl -\textgreater Ar, Pr, Rw & Pr -\textgreater Ar, Cl, Rw & Rw -\textgreater Ar, Cl, Pr & Average \\ \hline \hline
\multicolumn{6}{|c|}{\textbf{Teacher: ResNet50 --- Student: AlexNet}} \\ \hline
Source only  & 33.4                        & 37.6                        & 32.4                        & 39.3                        & 35.7    \\ \hline
DAN                                    & 39.7                        & 43.2                        & 39.4                        & 47.8                        & 42.5    \\ \hline
RevGrad                                & 42.1                        & 45.1                        & 41.1                        & 48.4                        & 44.2    \\ \hline
AMEANS                                 & 44.6                        & 47.6                        & 42.8                        & 50.2                        & 46.3    \\ \hline
MT-MTDA Mixed (ours)                                & 48.6                        & 48.1                       & 42.3                        & 53.0                       & 48.0   \\ \hline
MT-MTDA (ours)                                & \textbf{48.8}                        & \textbf{50.1}                        & \textbf{44.0}                        & \textbf{56.0}                        & \textbf{49.7}    \\ \hline \hline
\multicolumn{6}{|c|}{\textbf{Teacher: ResNext101 --- Student: ResNet50}} \\ \hline
Source only  & 47.6                        & 42.6                        & 44.2                        & 51.3                        & 46.4    \\ \hline
DAN                                    & 55.6                        & 56.6                        & 48.5                        & 56.7                        & 54.3    \\ \hline 
RevGrad                                & 58.4                        & 58.1                        & 52.9                        & 62.1                        & 57.9    \\ \hline
AMEANS                                 & 64.3                        & 65.5                        & 59.5                        & 66.7                        & 64.0    \\ \hline
MT-MTDA Mixed (ours)                                & \textbf{64.9}                        & 66.3                       & \textbf{60.2}                        & \textbf{66.9}                       & \textbf{64.6}   \\ \hline
MT-MTDA (ours)                                & 64.6                        & \textbf{67.1}                        & 59.0                        & 66.4                        & 64.3    \\ \hline
\end{tabular}
}
\end{table*}

From Tables \ref{tb:Digits5_comp_alexnet_weighted}, \ref{tb:office31_comp_weighted}, \ref{tb:officehome_comp_weighted}, our weighted results are still consistent with our equal-weight results in the main paper. Our method performs better than current state-of-the-art method in all cases except on Office31 with ResNet50. These results show that our method does not improve upon state-of-the-art by having a good accuracy on an easy case of domain adaptation with huge amount of data but it improves in more general manner.

\subsection{Additional Comparison on Each Target}

As mentioned in the main paper, we present more results on each separate target domain comparing to a standard STDA baseline \cite{GRL} on OfficeHome using AlexNet.

\begin{table*}[htbp]
\centering
\caption{Average accuracy of proposed and baseline STDA methods for individual and overall target datasets on OfficeHome dataset using AlexNet}
\label{tb:each_target_all_officeHome}
\resizebox{1.0\textwidth}{!}{
\begin{tabular}{|l|r|r|r|r|r|r|r|r|r|r|r|r|r|r|r|r|}
	\hline
	Alexnet & \multicolumn{4}{c|}{Ar $\xrightarrow{}$ Cl, Pr, Rw}                                                                & \multicolumn{4}{c|}{Cl $\xrightarrow{}$ Ar, Pr, Rw}                                                                & \multicolumn{4}{c|}{Pr $\xrightarrow{}$ Ar, Cl, Rw}                                                                & \multicolumn{4}{c|}{Rw $\xrightarrow{}$ Ar, Cl, Pr}                                                                            \\ 
	       & Ar $\xrightarrow{}$ Cl & Ar $\xrightarrow{}$ Pr & Ar $\xrightarrow{}$ Rw & Avg  & Cl $\xrightarrow{}$ Ar & Cl $\xrightarrow{}$ Pr & Cl $\xrightarrow{}$ Rw & Avg  & Pr $\xrightarrow{}$ Ar & Pr $\xrightarrow{}$ Cl & Pr $\xrightarrow{}$ Rw & Avg  & Rw $\xrightarrow{}$ Ar & Rw $\xrightarrow{}$ Cl & Rw $\xrightarrow{}$  Pr & Avg  \\ \hline \hline
	RevGrad STDA    & \textbf{36.4}                   & 45.2                   & 54.7                   & 45.4 & 35.2                   & 51.8                   & \textbf{55.1}                   & 47.4 & 31.6                   & \textbf{39.7}                   & \textbf{59.3}                   & \textbf{43.5} & 45.7                   & \textbf{46.4}                   & 65.9                    & 52.6 \\ \hline \hline
	AMEANS          & -                      & -                      & -                      & 44.6 & -                      & -                      & -                      & 45.6 & -                      & -                      & -                      & 41.4 & -                      & -                      & -                       & 49.3 \\ \hline
	MT-MTDA  & 34.1                   & \textbf{52.6}                   & \textbf{59.7}                   & \textbf{48.8} & \textbf{40.7}                   & \textbf{52.0}                   & 53.5                   & \textbf{48.7} & \textbf{36.5}                   & 33.7                   & 58.6                   & 42.9 & \textbf{55.0}                   & 42.0                   & \textbf{70.3}                    & \textbf{55.7} \\ \hline
\end{tabular}
}
\end{table*}

From Table \ref{tb:each_target_all_officeHome}, we can draw a similar conclusion of the main paper. Our method performs in average better than multiple STDA on different target domains. This shows that we can have one model handling different target domains without sacrificing computational power or memory.





\subsection{TSNE Visualization}
\begin{figure*}[!b]
    \centering
    \includegraphics[width=\textwidth]{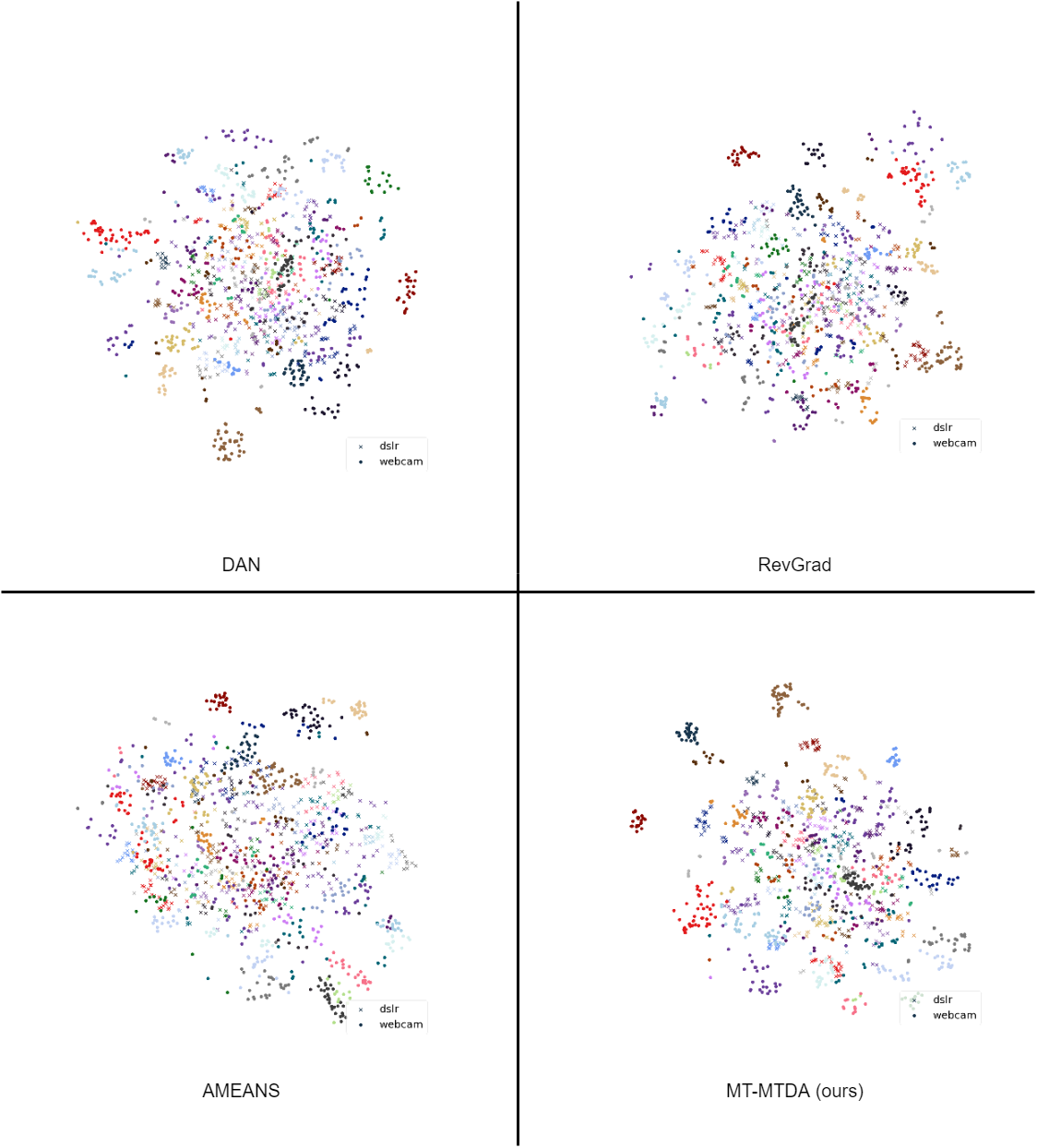}
    \caption{T-SNE visualization of all baselines methods versus MT-MTDA (ours)}
    \label{fig:TSNE_ALL}
\end{figure*}

In this section, we add the TSNE of RevGrad\cite{GRL} and DAN\cite{MMD_ICLR} and provide a higher resolution of the previous TSNE. From Figure \ref{fig:TSNE_ALL}, we can see that features between different target domains can be mixed together even when there's a blending mechanism like in \cite{BlendMTDA}.





\end{document}